\documentclass[10pt,twocolumn,letterpaper]{article}

\usepackage{iccv}
\usepackage{times}
\usepackage{epsfig}
\usepackage{graphicx}
\usepackage{amsmath}
\usepackage{amssymb}

\usepackage[pagebackref=true,breaklinks=true,letterpaper=true,colorlinks,bookmarks=false]{hyperref}
\usepackage[sort, numbers]{natbib}

\usepackage{enumitem}
\usepackage{tablefootnote}
\usepackage{multirow}
\usepackage{capt-of}
\usepackage{fixltx2e}
\usepackage{floatrow}
\usepackage{pifont}
\newcommand{\cmark}{\text{\ding{51}}}
\newcommand{\xmark}{\text{\ding{55}}}

 \iccvfinalcopy 

\def\iccvPaperID{1097} 

\newcommand{\x}{\times}

\newcommand{\Yh}{\hat{Y}}
\newcommand{\sub}{\textsubscript} 
\newcommand{\UCF}{UCF\textunderscore CROWD\textunderscore 50 }

\ificcvfinal\fi
\begin{document}

\title{Pushing the Frontiers of Unconstrained Crowd Counting: New Dataset and Benchmark Method}

\author{Vishwanath A. Sindagi \qquad Rajeev Yasarla \qquad Vishal M. Patel\\
	Department of Electrical and Computer Engineering,\\
	Johns Hopkins University, 3400 N. Charles St, Baltimore, MD 21218, USA\\
	{\tt\small \{vishwanathsindagi,rajeevyasarla,vpatel36\}@jhu.edu}
}
\maketitle
\thispagestyle{empty}

\begin{abstract}

In this work, we propose a novel crowd counting network that progressively generates crowd density maps via residual error estimation. The proposed method uses VGG16 as the backbone network and employs density map generated by the final layer as a coarse prediction to refine and generate finer density maps in a progressive fashion using residual learning. Additionally, the residual learning is guided by an uncertainty-based confidence weighting mechanism that permits the flow of only high-confidence residuals in the refinement path. The proposed Confidence Guided Deep Residual Counting Network (CG-DRCN) is evaluated on recent complex datasets, and it achieves significant improvements in errors.

Furthermore, we introduce a new large scale  unconstrained crowd counting dataset (JHU-CROWD) that is $\sim$2.8 $\times$ larger than the most recent crowd counting datasets in terms of the number of images.  It contains  4,250 images with 1.11 million annotations. In comparison to existing datasets, the proposed dataset is collected under a variety of  diverse scenarios and environmental conditions. Specifically, the dataset includes several images with weather-based degradations and illumination variations in addition to many distractor images, making it a very challenging dataset. Additionally, the dataset consists of rich annotations at both image-level and head-level.  Several recent methods are evaluated and compared on this dataset. 

\end{abstract}

\section{Introduction}

With burgeoning population and rapid urbanization, crowd gatherings have become more prominent in the recent years. Consequently, computer vision-based crowd analytics and surveillance \cite{li2015crowded,zhan2008crowd,idrees2013multi,zhang2015cross,zhang2016single,sindagi2017generating,sam2017switching,chan2008privacy,rodriguez2011density,zhu2014crowd,li2014anomaly,mahadevan2010anomaly, marsden2018people,sindagi2019dafe,sam2019almost}  have received increased interest. Furthermore, algorithms developed for the purpose of crowd analytics have found applications in other fields such as    agriculture monitoring  \cite{lu2017tasselnet}, microscopic biology \cite{lempitsky2010learning}, urban planning and environmental survey \cite{french2015convolutional,zhan2008crowd}. Current state-of-the-art counting networks achieve impressive error rates on a variety of datasets that contain numerous challenges. Their success can be broadly attributed to two major  factors: (i) design of novel convolutional neural network (CNN) architectures specifically for improving count performance \cite{zhang2015cross, walach2016learning, onoro2016towards, sam2017switching, sindagi2017cnnbased, ranjan2018iterative, cao2018scale,sam2018top}, and  (ii) development and publication of challenging datasets \cite{idrees2013multi,zhang2015cross,zhang2016single,idrees2018composition}. In this paper, we consider both of the above factors  in an attempt to further improve the crowd counting performance.

Design of novel networks specifically for the task of counting has improved the counting error by leaps and bounds. Architectures have evolved from the simple ones like \cite{zhang2015cross} which consisted of a set of convolutional and fully connected layers, to the most recent complex architectures like SA-Net \cite{cao2018scale} which consists of a set of scale aggregation modules. Typically, most existing works (\cite{zhang2015cross,zhang2016single,  walach2016learning, onoro2016towards, sam2017switching, sindagi2017cnnbased, ranjan2018iterative, cao2018scale,babu2018divide, sindagi2017generating, cao2018scale,sindagi2019ha}) have designed their networks by laying  a strong emphasis on addressing large variations of scale in crowd images. While this strategy of developing robustness towards scale changes has resulted in significant performance gains, it is nevertheless important to exploit other properties like in \cite{ranjan2018iterative, shen2018adversarial,shi2018crowd_negative} to further the improvements.

In a similar attempt, we exploit residual learning mechanism for the purpose of improving crowd counting. Specifically, we present a novel  design based on the VGG16 network \cite{simonyan2014very}, which employs residual learning to  progressively  generate better quality crowd density maps.  This use of residual learning is inspired by its success in several other tasks like  super-resolution \cite{tai2017image,kim2016accurate,lim2017enhanced,ke2017srn,kim2016accurate}. Although this technique results in improvements in performance, it is important to ensure that only highly confident residuals are used in order to ensure  the effectiveness of residual learning.  To address this issue, we draw inspiration from  the success of uncertainty-based learning mechanism \cite{kendall2017uncertainties,zhu2017deep,devries2018learning}. We propose an uncertainty-based confidence weighting module  that captures high-confidence regions in the feature maps to focus on during the residual learning.  The confidence weights ensure that only highly confident residuals get propagated to the output, thereby increasing the effectiveness of the residual learning mechanism.  

In addition to the new network design, we identify the next set of challenges that require attention from the crowd counting research community and collect a large-scale dataset collected under a variety of conditions.  Existing efforts like \UCF \cite{idrees2013multi}, World Expo '10 \cite{zhang2015cross} and ShanghaiTech \cite{zhang2016data} have progressively increased the complexity of the datasets in terms of average count per image, image diversity \etc. While these datasets have enabled rapid progress in the counting task, they suffer from  shortcomings such as limited number of training samples, limited diversity in terms of environmental conditions, dataset bias in terms of positive samples, and limited set of annotations. More recently, Idrees \etal \cite{idrees2018composition} proposed a new dataset called UCF-QNRF that alleviates some of these challenges. Nevertheless, they do not specifically consider some of the challenges such as adverse environmental conditions, dataset bias and limited annotation data.

To address these issues, we propose a new large-scale  unconstrained dataset with a total of 4,250 images (containing 1,114,785 head annotations) that are collected under a variety of conditions.  Specific care is taken to include images captured under various weather-based degradations. Additionally, we include a set of distractor images that are similar to the crowd images but do not contain any crowd.  Furthermore, the  dataset also provides a much richer set of annotations at both image-level and head-level. We also benchmark several representative counting networks, providing an overview of the state-of-the-art performance.  

Following are our key contributions in this paper:
\begin{itemize}[topsep=0pt,noitemsep,leftmargin=*]
	\item We propose a  crowd counting network that progressively incorporates residual mechanism to estimate high quality density maps.  Furthermore, a set of uncertainty-based confidence weighting modules are introduced in the network  to improve the efficacy of  residual learning. 
	\item  We propose a new large-scale unconstrained crowd counting dataset with the largest number of images till date. The dataset specifically includes a number of images collected under adverse weather conditions. Furthermore, this is the first counting dataset that provides a rich set of annotations such as occlusion, blur, image-level labels,  \etc. 
\end{itemize}

\section{Related work}
\label{sec:related}

\noindent\textbf{Crowd Counting.}  Traditional approaches for crowd counting from single images are based on hand-crafted representations and different regression techniques. Loy \etal \cite{loy2013crowd} categorized these methods into (1) detection-based methods \cite{li2008estimating} (2) regression-based methods \cite{ryan2009crowd,chen2012feature,idrees2013multi} and (3) density estimation-based methods \cite{lempitsky2010learning,pham2015count,xu2016crowd}. Interested readers are referred to \cite{chen2012feature,li2015crowded} for  more comprehensive study of different crowd counting methods.

\begin{figure*}[ht!]
	\begin{center}
		\includegraphics[width=0.85\linewidth]{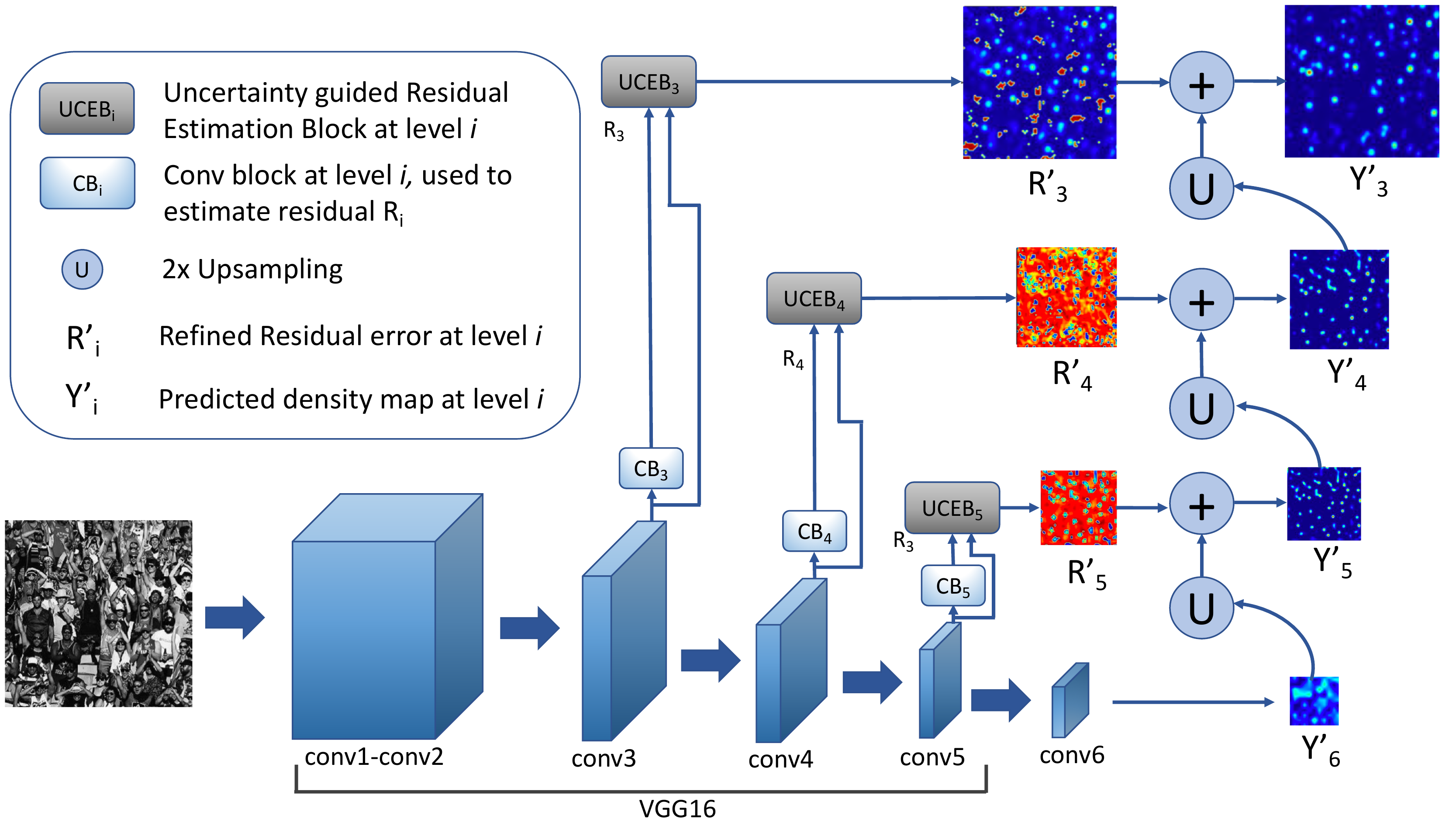}
	\end{center}
	\vskip -15pt \caption{Overview of the proposed method. Coarse density map from the deepest layer of the base network is refined using the residual map estimated by the shallower layer. The residual estimation is performed by convolutional block, $CB_i$ and is further refined in $UCEB_i$. Note that, the conv features from the main branch are first reduced to 32 dimensions using 1$\times$1 conv before forwarding them to $UCEB_i$ along with $R_i$.  In the residual maps, red indicates negative values and cyan indicates positive value.}
	\label{fig:arch}
\end{figure*}

Recent advances in CNNs have been exploited for the task of crowd counting and these methods \cite{wang2015deep,zhang2015cross,sam2017switching,arteta2016counting,walach2016learning,onoro2016towards,zhang2016single,sam2017switching,sindagi2017generating,boominathan2016crowdnet,wang2018defense,onoro2018learning} have demonstrated significant improvements over the traditional methods. A recent survey \cite{sindagi2017survey} categorizes these approaches based on the network property and the inference process. Walach \etal \cite{walach2016learning} used CNNs with layered boosting approach to learn a non-linear function between an image patch and count. Recent work \cite{zhang2016single,onoro2016towards} addressed the scale issue using different architectures. Sam \etal \cite{sam2017switching} proposed a VGG16-based  switching classifier that first identifies appropriate regressor based on the content of the input image patch. More recently, Sindagi \etal \cite{sindagi2017generating} proposed to incorporate global and local context from the input image into the density estimation network. In another approach, Cao \etal \cite{cao2018scale} proposed a encoder-decoder network with scale aggregation modules.

In contrast to these  methods that emphasize on specifically addressing large-scale variations in head sizes, the most recent methods (\cite{babu2018divide} ,\cite{shen2018adversarial}, \cite{shi2018crowd_negative}, \cite{liu2018leveraging}, \cite{ranjan2018iterative}) have focused on other properties of the problem. For instance, 
Babu \etal \cite{babu2018divide} proposed a mechanism  to incrementally increase the network capacity  conditioned on the dataset. Shen \etal \cite{shen2018adversarial} overcame the issue of blurred density maps by utilizing  adversarial loss.   In a more recent approach, Ranjan \etal \cite{ranjan2018iterative} proposed a two-branch network to  estimate density map in a cascaded manner.  Shi \etal \cite{shi2018crowd_negative} employed deep negative correlation based learning for more generalizable features. Liu \etal \cite{liu2018leveraging} used  unlabeled data for counting by proposing a new framework that involves learning to rank. 

Recent approaches like \cite{liu2018adcrowdnet,wan2019residual,zhao2019leveraging,sindagi2019inverse,sindagi2019ha} have aimed at incorporating various forms of related information like attention \cite{liu2018adcrowdnet}, semantic priors \cite{wan2019residual}, segmentation \cite{zhao2019leveraging}, inverse attention \cite{sindagi2019inverse}, and hierarchical attention \cite{sindagi2019ha} respectively into the network.   Other techniques such as \cite{jiang2019crowd,shi2019revisiting,liu2019context, zhang2019wide} leverage features from different layers of the network using different techniques like   trellis style encoder decoder \cite{jiang2019crowd},  explicitly considering perspective \cite{shi2019revisiting},  context information  \cite{liu2019context},  and multiple views \cite{zhang2019wide}.

\noindent\textbf{Crowd Datasets.}  Crowd counting datasets have evolved over time with respect to a number of factors such as size, crowd densities, image resolution, and diversity.  UCSD \cite{chan2008privacy} is among one of the early datasets proposed for counting and it contains 2000 video frames of low resolution with 49,885 annotations. The video frames are collected from a single frame and typically contain low density crowds. Zhang \etal \cite{zhang2015cross} addressed the limitations of UCSD dataset by introducing  the WorldExpo dataset that contains 108 videos with a total of 3,980 frames belonging to 5 different scenes. While the UCSD and WorldExpo datasets contain only low/low-medium densities, Idrees \etal \cite{idrees2013multi} proposed the \UCF dataset specifically for very high density crowd scenarios. However, the dataset consists of only 50 images rendering it impractical for training deep networks. Zhang \etal \cite{zhang2016single} introduced the ShanghaiTech dataset which has better diversity in terms of scenes and density levels as compared to earlier datasets.  The dataset is split into two parts: Part A (containing high density crowd images) and Part B (containing low density crowd images). The entire dataset contains 1,198 images with 330,165 annotations. Recently, Idrees \etal \cite{idrees2018composition} proposed  a new large-scale crowd dataset containing 1,535 high density images images with a total of  1.25 million annotations.  Wang \etal \cite{wang2019learning} introduced a synthetic crowd dataset  that contains diverse scenes. In addition, they  proposed  a SSIM based CycleGAN \cite{zhu2017unpaired} for adapting the network trained on   synthetic images to real world images. \\

\begin{figure*}[t!]
	\begin{center}
		\includegraphics[width=0.195\linewidth]{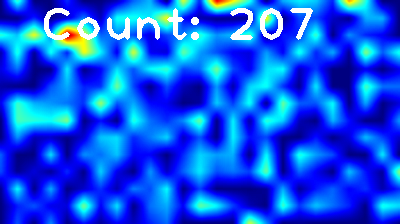}
		\includegraphics[width=0.195\linewidth]{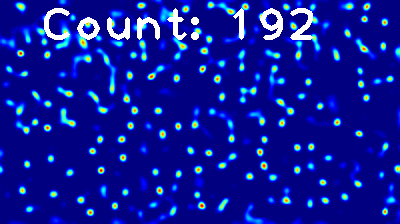}
		\includegraphics[width=0.195\linewidth]{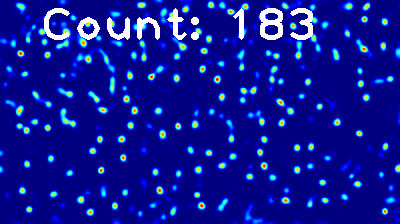}
		\includegraphics[width=0.195\linewidth]{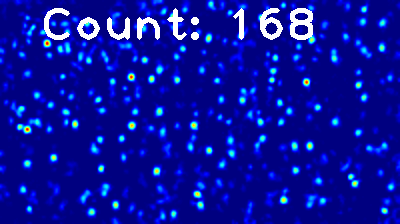}
		\includegraphics[width=0.195\linewidth]{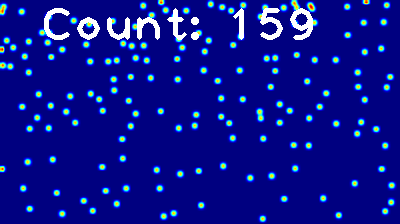}\\
		(a)\hskip 90pt(b)\hskip 90pt(c)\hskip 90pt(d)\hskip 90pt(e)  
	\end{center}
	\vskip -18pt \caption{Density maps estimated by different layers of the proposed network. (a) $\Yh_6$ (b) $\Yh_5$  (c) $\Yh_4$  (d) $\Yh_3$   (e) $Y $(ground-truth). It can be observed that the output of the deepest layer ($\Yh_6$) looks very coarse, and it is refined in a progressive manner using the residual learned by the conv blocks $CB_5,CB_4,CB_3$ to obtain the $\Yh_5,\Yh_4, \Yh_3$ respectively. Note that fine details and the total count in the density maps improve as we  move from $\Yh_6$  to $\Yh_3$.}
	\label{fig:coarse_to_fine}
\end{figure*}

\section{Proposed method}

In this section, we present the details  of  the proposed Confidence Guided Deep Residual Crowd Counting (CG-DRCN)  along with the training and inference specifics. Fig. \ref{fig:arch} shows the architecture of the proposed network. 

\subsection{Base network}
\label{ssec:base_network}
Following recent approaches \cite{sindagi2017generating,sam2017switching,cao2018scale}, we perform counting based on the density estimation framework. In this framework, the network is trained to estimate the density map ($\hat{Y}$) from an input crowd image ($X$). The target density map  ($Y$) for training the network is generated by imposing normalized 2D Gaussian at head locations provided by the dataset annotations:  $Y(x) = \sum_{{x_g \in S}}\mathcal{N}(x-x_g,\sigma)$,
where, $S$ is the set of all head locations ($x_g$) in the input image and $\sigma$ is scale parameter of 2D Gaussian kernel. Due to this formulation, the density map contains per-pixel density information of the scene, which when integrated results in the  count of people in the image.

The proposed network consists of conv1$\sim$conv5  layers ($C_1-C_5$) of the VGG16 architecture as a part of the backbone, followed by a conv block ($CB_6$) and a max-pooling layer with stride 2.  First, the input image (of size $W\x H$) is passed through $C_1-C_5$, $CB_6$ and the max pooling layer to produce the corresponding density map ($\Yh _6$) of size $\frac{W}{32}\x \frac{H}{32}$. $CB_6$ is defined by  \textit\{conv\sub{512,32,1}-relu-conv\sub{32,32,3}-relu-conv\sub{32,1,3}\}\footnote{ \label{fn:conv}\textit{conv\sub{N\sub{i},N\sub{o},k}} denotes conv layer (with \textit{N\sub{i}} input channels, \textit{N\sub{o}} output channels, \textit{k}$\times$\textit{k} filter size), \textit{relu} denotes ReLU activation}). Due to its low resolution, ($\Yh _6$)  can be considered as a coarse estimation, and learning this will implicitly incorporate global context in the image due the large receptive field at the deepest layer in the network.

\subsection{Residual learning}
\label{ssec:residual_learning}
Although $\Yh _6$ provides a good estimate of the number of people in the image, the density map lacks several local details as shown in Fig. \ref{fig:coarse_to_fine} (a).  This is because deeper layers learn to capture abstract concepts and tend to lose low level details in the image. On the other hand, the shallower layers have relatively more detailed local information as compared to their deeper counterparts \cite{ranjan2016hyperface}. Based on this observation, we  propose to refine the coarser density maps by employing shallower layers in a residual learning framework. This refinement mechanism is inspired in part by several leading work on super-resolution \cite{tai2017image,kim2016accurate,lim2017enhanced} that incorporate residual learning to learn finer details required to generate a high quality  super-resolved image. Specifically, features from $C_5$ are forwarded through a conv-block ($CB_5$) to generate a residual map $R_5$, which is then added to an appropriately up-sampled version of $\Yh_6$ to produce the density map $\Yh_5$ of size $\frac{W}{16}\x \frac{H}{16}$, \ie, 
\begin{equation}
\Yh_5 = R_5 + up(\Yh_6).
\end{equation}
Here, $up()$ denotes up-sampling by a factor of 2$\x$ via bilinear interpolation. By enforcing $CB_5$ to learn a residual map, the network  focuses on the local errors emanating from  the deeper layer, resulting in better learning of the offsets required to refined the coarser density map. $CB_5$ is defined by  \textit\{conv\sub{512,32,1}-relu-conv\sub{32,32,3}-relu-conv\sub{32,1,3}\}\textsuperscript{\ref{fn:conv}}.

The above refinement is further repeated to recursively generate  finer density maps $\Yh_4$ and $\Yh_3$ using the feature maps from the shallower layers $C_4$ and $C_3$, respectively. Specifically, the output of  $C_4$ and $C_3$ are forwarded through $CB_4,CB_3$  to learn residual maps $R_4$ and $R_3$, which are then added to the appropriately up-sampled versions of the coarser maps $\Yh_5$ and $\Yh_4$ to produce $\Yh_4$ and $\Yh_3$ respectively in that order. $CB_4$ is defined by  \textit\{conv\sub{512,32,1}-relu-conv\sub{32,32,3}-relu-conv\sub{32,1,3}\}\textsuperscript{\ref{fn:conv}}. $CB_3$ is defined by  \textit\{conv\sub{256,32,1}-relu-conv\sub{32,32,3}-relu-conv\sub{32,1,3}\}\textsuperscript{\ref{fn:conv}}. Specifically, $\Yh_4$ and $\Yh_3$ are obtained as follows: $\Yh_4 = R_4 + up(\Yh_5), \Yh_3 = R_3 + up(\Yh_4)$.



\subsection{Confidence guided residual learning}
\label{ssec:ceb}

In order to improve the efficacy of the residual learning mechanism discussed above, we propose an uncertainty guided confidence estimation  block (UCEB) to guide the refinement process. The task of conv blocks $CB_5, CB_4, CB_3$ is to capture residual errors that can be incorporated into the coarser density maps to produce high quality density maps in the end. For this purpose, these conv blocks employ feature maps from shallower conv layers $C_5,C_4,C_3$.
Since these conv layers  primarily trained for estimating the coarsest density map, their features have high responses in regions where crowd is present, and hence, they may  not necessarily produce effective residuals. In order to overcome this issue, we propose to gate the residuals that are not effective using uncertainty estimation. Inspired by uncertainty estimation in CNNs \cite{kendall2017uncertainties,zhu2017deep,devries2018learning,Yasarla_2019_CVPR}, we aim to model pixel-wise aleatoric uncertainty of the residuals  estimated by $CB_5, CB_4, CB_3$.  That is we, predict the pixel-wise confidence (inverse of the uncertainties) of the residuals which are then     used to gate the residuals before being passed on to the subsequent outputs. This ensures that only highly confident residuals get propagated to the output.

In terms of the overall architecture, we introduce a set of UCEBs as shown in Fig. \ref{fig:arch}. Each residual branch consists of one such block. The UCEB$_i$ takes the residual $R_i$ and dimensionality reduced features from the main branch  as input, concatenates them, and forwards it through a set of conv layers (\textit\{conv\sub{33,32,1}-relu-conv\sub{32,16,3}-relu-conv\sub{16,16,3}-relu-conv\sub{16,1,1}\}) and produces a confidence map $CM_i$ which is then multiplied element-wise with the input to form the refined residual map: $\hat{R}_i = R_i\odot CM_i$. Here  $\odot$ denotes element-wise multiplication.  

In order to learn these confidence maps, the loss function $L_f$ used to train the network is defined as follows,
\begin{equation}
\label{eq:lossnew}
L_f = L_d - \lambda_c L_c,
\end{equation} 
where, $\lambda_{c}$ is a regularization constant, $L_d$ is the pixel-wise regression loss to minimize the density map prediction error and is defined as:
\begin{equation}
\label{eq:lossdensity}
L_d = \sum_{i\in\{3,4,5,6\}}\| (CM_i \odot Y_i) - (CM_i \odot \Yh_i)\|_2,
\end{equation}
where,  $\Yh_i$ is the predicted density map, $i$ indicates the index of the conv layer from which the predicted density map is taken,  $Y_i$ is the corresponding target. 

$L_c$ is the confidence guiding loss, defined as,
\begin{equation}
L_c = \sum_{i\in\{3,4,5,6\}} \sum_{j=1}^{H} \sum_{k=1}^{W}  \log(CM_i^{j,k}), 
\end{equation} 
where, $W\times H$ is the dimension of the confidence map $CM_i$. As it can be seen from  Eq. \eqref{eq:lossnew}, the loss $L_f$ has two parts $L_d$ and $L_c$. The first term minimizes the Euclidean distance between the prediction and target features, whereas $L_c$ maximizes the confidence scores $CM_i$ by making them closer to 1. 

Fig. \ref{fig:coarse_to_fine} illustrates the output density maps ($\Yh_6,\Yh_5,\Yh_4,\Yh_3$) generated by the proposed network for a sample crowd image. It can be observed that the density maps progressively improve in terms of fine details and the count value.

\subsection{Training and inference details}
\label{ssec:training_details}
\noindent The training dataset is obtained by cropping patches from multiple random locations in each training image. The cropped patch-size is 224$\times$224. We randomly sub-sample 10\% of the training set (before cropping) and keep it aside for validating the training models. We use the Adam optimizer to train the network. We use a learning rate of 0.00001 and a momentum of 0.9.


For inference, the density map $\Yh_3$ is considered as the final output. 
The count performance is measured using the standard error metrics: mean absolute error ($MAE$) and mean squared error ($MSE$). These metrics are defined   as follows: $MAE = \frac{1}{N}\sum_{i=1}^{N}|Y_i-Y'_i|$ and  $MSE = \sqrt{\frac{1}{N}\sum_{i=1}^{N}|Y_i-Y'_i|^2}$ respectively, where  $N$ is the number of test samples, $Y_i$ is the ground-truth count and $Y'_i$ is the estimated count corresponding to the $i^{th}$ sample.

\section{JHU-CROWD: Unconstrained Crowd Counting Dataset}
In this section, we first motivate the need for a new crowd counting dataset, followed by a detailed description of the various factors and conditions while collecting the dataset. 

\begin{table}[b!]
	\centering
	\caption{Summary of images collected under adverse conditions.}
	\label{tab:weather}
	\resizebox{0.9\textwidth}{!}{%
		\begin{tabular}{|l|c|c|c|c|}
			\hline
			\textbf{Degradation type} & \textbf{Rain} & \textbf{Snow} & \textbf{Fog/Haze} & \textbf{Total} \\ \hline
			Num. of images            & 151           & 190           & 175               & 516            \\   
			Num. of annotations       & 32,832        & 32,659        & 37,070            & 102,561         \\ \hline
		\end{tabular}%
	}
	
\end{table}

\begin{figure*}[ht!]
	\begin{center}
		\includegraphics[width=0.19\linewidth, height=0.13\linewidth]{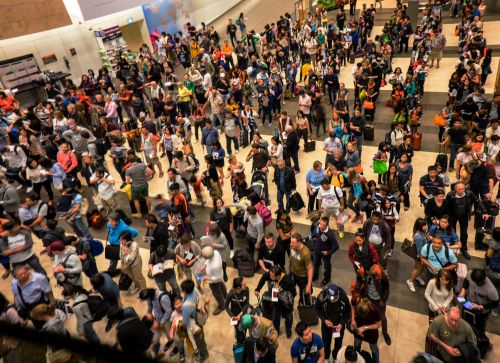}
		\includegraphics[width=0.19\linewidth, height=0.13\linewidth]{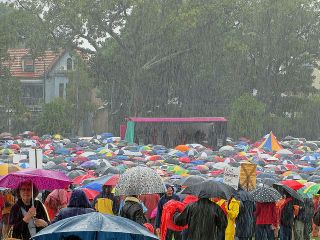}
		\includegraphics[width=0.19\linewidth, height=0.13\linewidth]{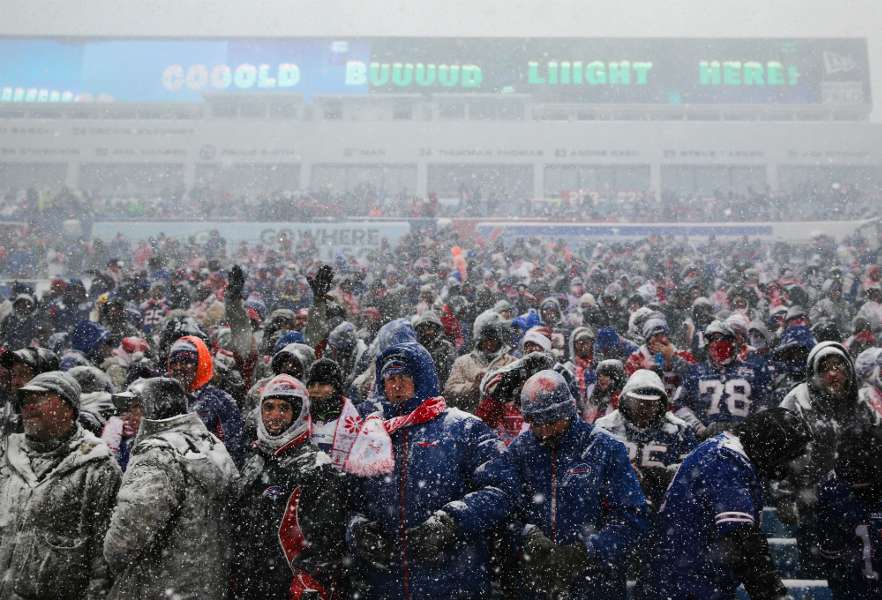}
		\includegraphics[width=0.19\linewidth, height=0.13\linewidth]{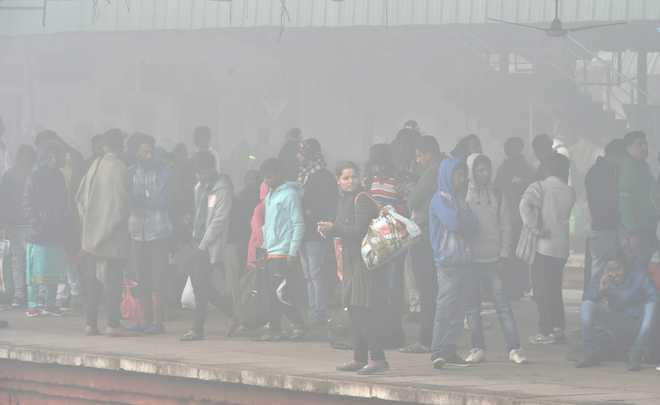}
		\includegraphics[width=0.19\linewidth, height=0.13\linewidth]{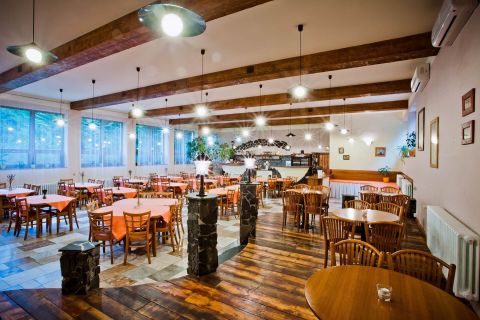}
		
		\includegraphics[width=0.19\linewidth, height=0.13\linewidth]{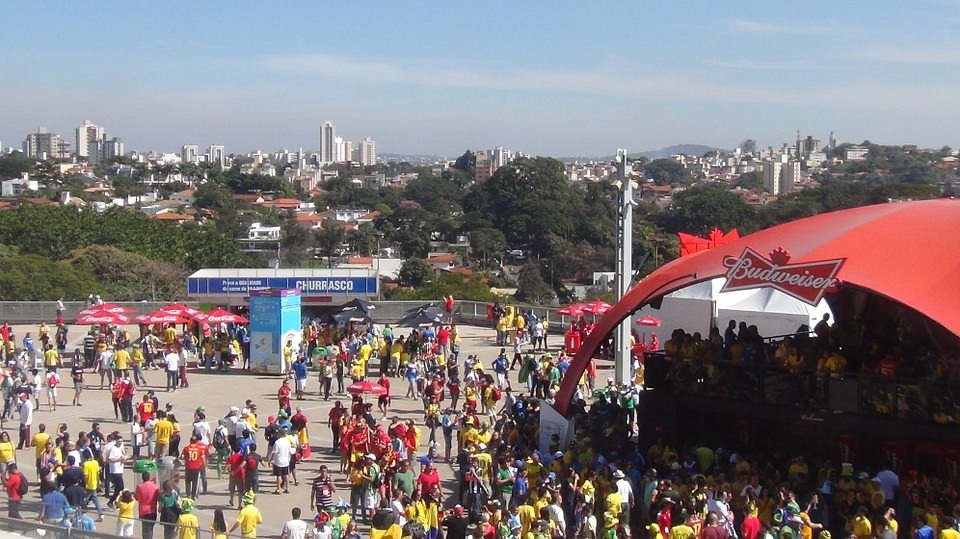}
		\includegraphics[width=0.19\linewidth, height=0.13\linewidth]{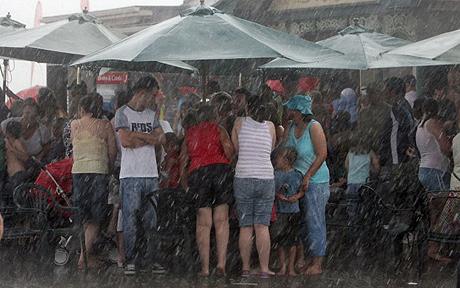}
		\includegraphics[width=0.19\linewidth, height=0.13\linewidth]{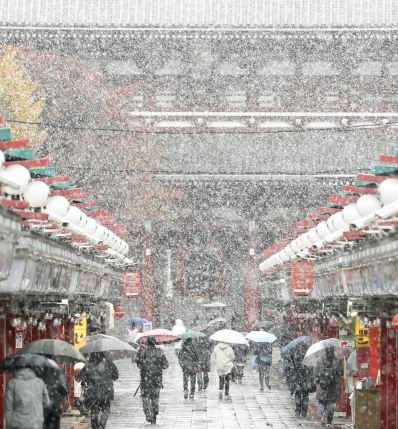}
		\includegraphics[width=0.19\linewidth, height=0.13\linewidth]{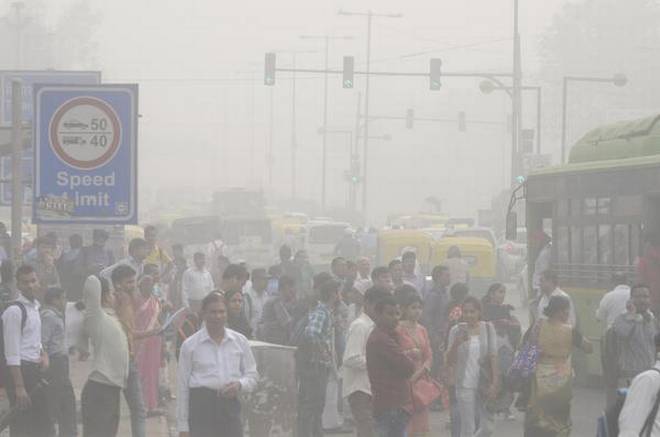}
		\includegraphics[width=0.19\linewidth, height=0.13\linewidth]{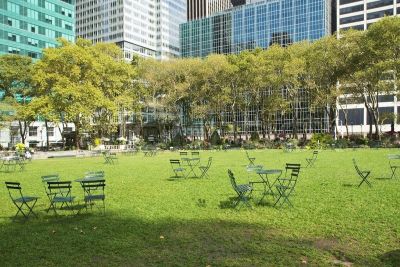}\\
		\includegraphics[width=0.19\linewidth, height=0.13\linewidth]{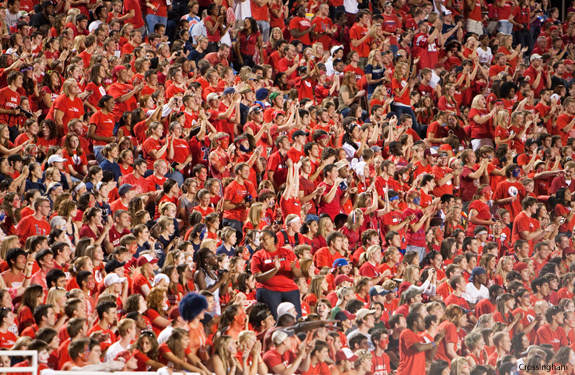}
		\includegraphics[width=0.19\linewidth, height=0.13\linewidth]{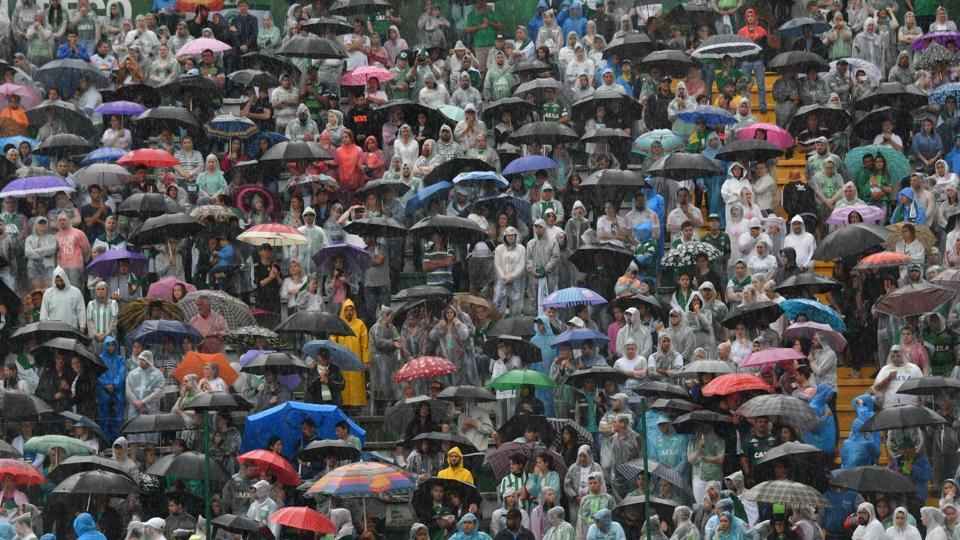}
		\includegraphics[width=0.19\linewidth, height=0.13\linewidth]{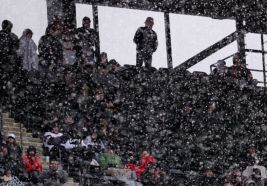}
		\includegraphics[width=0.19\linewidth, height=0.13\linewidth]{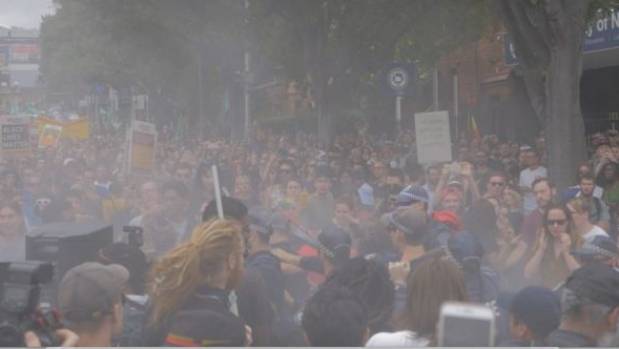}
		\includegraphics[width=0.19\linewidth, height=0.13\linewidth]{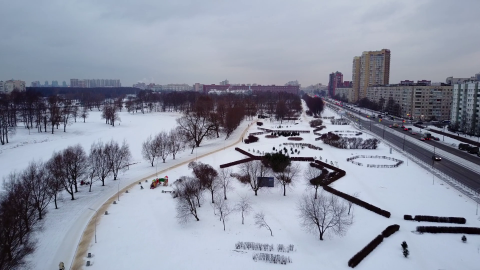}\\
		(a)\hskip 85pt(b)\hskip 85pt(c)\hskip 85pt(d)\hskip 85pt(e)  
	\end{center}
	\vskip -10pt \caption{Representative samples of the images in the JHU-CROWD dataset. (a) Overall (b) Rain (c) Snow (d) Haze (e) Distractors.}
	\label{fig:dataset_samples}
\end{figure*}

\begin{table*}[ht!]
	\caption{Comparison of different datasets. P: Point-wise annotations for head locations, O: Occlusion level per head, B: Blur level per head, S: Size indicator per head, I: Image level labels.}
	\label{tab:dataset_summary}
	\resizebox{1\columnwidth}{!}{%
		\begin{tabular}{|l|c|c|c|c|c|c|c|c|}
			\hline
			\textbf{Dataset}      & \textbf{\begin{tabular}[c]{@{}c@{}}Num of\\ Images\end{tabular}} & \textbf{\begin{tabular}[c]{@{}c@{}}Num of\\ Annotations\end{tabular}} & \textbf{\begin{tabular}[c]{@{}c@{}}Avg\\  Count\end{tabular}} & \textbf{\begin{tabular}[c]{@{}c@{}}Max \\ Count\end{tabular}} & \textbf{\begin{tabular}[c]{@{}c@{}}Avg \\ Resolution\end{tabular}} & \textbf{\begin{tabular}[c]{@{}c@{}}Weather\\ degradations\end{tabular}} & \textbf{Distractors} & \textbf{\begin{tabular}[c]{@{}c@{}}Type of \\ annotations\end{tabular}} \\ \hline
			UCSD \cite{chan2008privacy}          & 2000 & 49,885  & 25 & 46  & 158$\times$238     & $\xmark$   & $\xmark$             & P     \\  
			Mall \cite{chen2012feature}          & 2000 & 62,325    & -      & 53     & 320$\times$240  & $\xmark$  & $\xmark$ & P    \\  
			\UCF \cite{idrees2013multi}          & 50     & 63,974      & 1279        & 4543       & 2101$\times$2888        & $\xmark$    & $\xmark$ & P  \\  
			WorldExpo '10 \cite{zhang2015cross} & 3980  & 199,923      & 50         & 253       & 576$\times$720 & $\xmark$   & $\xmark$ & P \\  
			ShanghaiTech \cite{zhang2016single}         & 1198    & 330,165  & 275 & 3139 & 598$\times$868 & $\xmark$ & $\xmark$ & P\\  
			UCF-QNRF \cite{idrees2018composition} & 1535 & 1,251,642 & 815   & 12865  & 2013$\times$2902    & $\xmark$ & $\xmark$  &P \\  
			\bf{JHU-CROWD (proposed)}       & 4250  & 1,114,785 & 262 & 7286 & 1450$\times$900 & $\cmark$ & $\cmark$ & P, O, B, S, I \\ \hline
		\end{tabular}
	}
	\vskip-10pt
\end{table*}

\subsection{Motivation and dataset details}
\label{ssec:motivation_dataset}
As discussed earlier, existing datasets (such as \UCF \cite{idrees2013multi}, World Expo '10 \cite{zhang2015cross} and ShanghaiTech \cite{zhang2016data})  have enabled researchers to develop novel counting networks that are robust to several factors such as variations in scale, pose, view \etc. Several recent methods have specifically addressed the large variations in scale by proposing different approaches such as multi-column networks \cite{zhang2016single}, incorporating global and local context \cite{sindagi2017generating}, scale aggregation network \cite{cao2018scale}, \etc. These methods are largely successful in addressing issues in the existing datasets, and there is pressing need to identify newer set of challenges that require attention from the crowd counting community. 

In what follows, we describe the shortcomings of existing datasets and discuss the ways in which we overcome them: 

\noindent(i) \textit{Limited number of training samples}:  Typically, crowd counting datasets have limited number of images available for training and testing. For example, ShanghaiTech dataset \cite{zhang2016single} has only 1,198 images and this low number of images results in lower diversity of the training samples. Due to this issue, networks trained on this dataset will have reduced generalization capabilities. Although datasets like Mall \cite{chen2012feature}, WorldExpo '10 \cite{zhang2015cross} have higher number of images, it is   important to note that these images are from a set of video sequences from surveillance cameras and hence, they have limited diversity in terms of background scenes and number of people. Most recently, Idrees \etal \cite{idrees2018composition} addressed this issue by introducing a high-quality dataset (UCF-QNRF) that has images collected from various geographical locations under a variety of conditions and scenarios. Although it has a large set of diverse scenarios, the number of samples is still limited from the perspective of training deep neural networks. 

To address this issue, we collect a new large scale  unconstrained dataset with a total of 4,250 images that are collected under a variety of conditions. Such a large number of images results in increased diversity in terms of count, background regions, scenarios \etc as compared to existing datasets.  The images are collected from several sources on the Internet using different keywords such as crowd, crowd+marathon, crowd+walking, crowd+India, \etc.

\noindent(ii) \textit{Absence of adverse conditions}: Typical application of crowd counting is  video surveillance in outdoor scenarios which involve regular weather-based degradations such as haze, snow, rain \etc. It is crucial that networks, deployed under such conditions, achieve more than satisfactory performance. 

To overcome this issue, specific care is taken during our dataset collection efforts to include images captured under various weather-based degradations such as rain, haze, snow, \etc (as as shown in Fig. \ref{fig:dataset_samples}(b-d) ).   Table \ref{tab:weather} summarizes images collected under adverse conditions.

\noindent(iii) \textit{Dataset bias}: Existing datasets focus on collecting only images with crowd, due to which a deep network trained on such a dataset may end up learning bias in the dataset. Due to this error, the network will erroneously predict crowd even in scenes that do not contain crowd.

In order to address this, we include a set of distractor images that are similar to crowd images but do not contain any crowd. These images can enable the network to avoid learning bias in the dataset. The total number of distractor images in the dataset is 100.  Fig \ref{fig:dataset_samples}(e) shows sample distractor images. 

\noindent(iv) \textit{Limited annotations}: Typically, crowd counting datasets provide point-wise annotations for every head/person in the image, \ie, each image is provided with a list of $x,y$ locations of the head centers. While these annotations enable the networks to learn the counting task,  absence of more information such as occlusion level, head sizes, blur level \etc limits the learning ability of the networks. For instance, due to the presence of large variations in perspective, size of the head is crucial to determine the precise count. One of the reasons for these missing annotations is that crowd images typically contain several people and it is highly labor intensive to obtain detailed annotations such as size. 

To enable more effective learning, we collect a much  richer set of annotations at both image-level and head-level. Head-level annotation include $x,y$ locations of heads and corresponding occlusion level, blur level and size level. Occlusion label has three levels: \{\textit{un-occluded, partially occluded, fully occluded}\}. Blur level has two labels: \{\textit{blur, no-blur}\}. Since obtaining the size is a much harder issue, each head is labeled with a size indicator. Annotators were instructed to first annotate the largest and smallest head in the image with a bounding box. The annotators were then instructed to assign a size level to every head in the image such that this size level is indicative of the relative size with respect to the smallest and largest annotated bounding box. Image level annotations include labels (such as \textit{marathon, mall, walking, stadium \etc }) and the weather conditions under which the images were captured. The total number of point-level annotations in the dataset are 1,114,785. 

\subsection{Summary and evaluation protocol}
Fig. \ref{fig:dataset_samples} illustrates representative samples of the images in the JHU-CROWD dataset under various categories.  Table \ref{tab:dataset_summary} summarizes the proposed JHU-CROWD dataset in comparison with  the existing ones. It can be observed that the proposed dataset is the largest till date in terms of the number of images and enjoys a host of other properties such as a richer set of  annotations, weather-based degradations and distractor images. With these properties, the proposed dataset will serve as a good complementary to other datasets such as UCF-QNRF.  
The dataset is randomly split into training and test sets, which contain 3,188 and 1,062 images respectively.

\begin{figure*}[ht!]
	\begin{center}
	\includegraphics[width=1\linewidth]{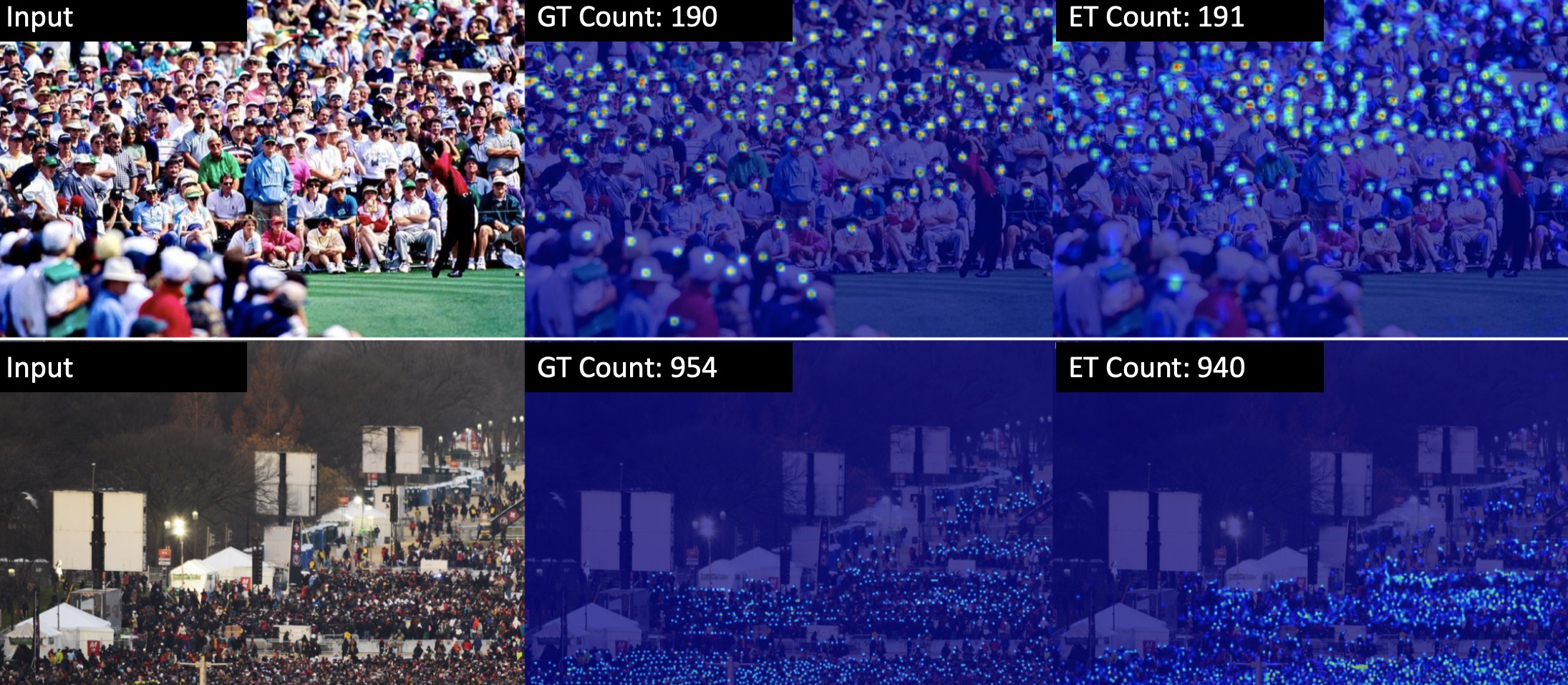}\\
		(a) \hskip 160pt (b) \hskip 160pt (c)
	\end{center}
	\vskip -18pt \caption{Results of the proposed dataset on sample images from the JHU-CROWD dataset. (a) Input image (b) Ground-truth density map (c) Estimated density map.}
	\label{fig:results_uccd}
\end{figure*}

Following the existing work, we perform evaluation using the  standard  MAE and MSE metrics.  Furthermore, these metrics are calculated for the following sub-categories of images: (i) Low density: images containing count between 0 and 50,  (ii) Medium density: images containing count between 51 and 500,  (iii) High density: images with count more than 500 people,  (iv) Distractors: images containing 0 count, (v) Weather-based degradations, and (vi) Overall. The metrics under these sub-categories will enable  a deeper understanding of the network performance.

\section{Experimental details and results}

In this section, we first discuss  the results of an ablation study conducted to analyze the effect of different components in the proposed network. This is followed by a discussion on benchmarking of recent crowd counting algorithms including the proposed residual-based counting network on the JHU-CROWD dataset.  Finally, we compared the  proposed method with recent approaches on the  ShanghaiTech \cite{zhang2016single} and UCF-QNRF \cite{idrees2018composition} datasets.

\subsection{Ablative Study}


Due to the presence of various complexities  such as high density crowds, large variations in scales, presence of occlusion, etc, we chose to perform the ablation study on JHU-CROWD dataset. 

The ablation study consisted of evaluating the following configurations of the  proposed method: (i) Base network: VGG16 network with an additional conv block ($CB_6$) at the end, (ii) Base network + R: the base network with residual learning as discussed in Section \ref{ssec:residual_learning} , (iii) Base network + R + UCEB ($\lambda_c=0$): the base network with residual learning guided by the confidence estimation blocks as discussed in Section \ref{ssec:ceb}. In this configuration, we aim to measure the performance due to the addition of the confidence estimation blocks without the uncertainty estimation mechanism by setting  $\lambda_c$ is set to $0$, (iv) Base network + R + UCEB ($\lambda_c=1$): the base network with residual learning guided by the confidence estimation blocks as discussed in Section \ref{ssec:ceb}. The results of these experiments are shown in Table \ref{tab:ablation}. It can be seen that  there are considerable improvements in the performance due to the inclusion of residual learning into the network. The use of confidence-based weighting of the residuals  results in further improvements, thus highlighting its significance in improving the efficacy of uncertainty-based residual learning.

\begin{table}[h!]
	\centering
	\caption{Results of ablation study on the JHU-CROWD dataset.}
	\label{tab:ablation}
	\resizebox{0.75\linewidth}{!}{
		\begin{tabular}{|l|c|c|}
			\hline
			Method & MAE & MSE \\			\hline
			Base network & 81.1 & 248.5 \\
			Base network + R  & 76.4 & 218.6 \\
			Base network + R + UCEB ($\lambda_c=0$)    & 74.6 & 215.5 \\
			Base network + R + UCEB ($\lambda_c=1$)   & 66.1 & 195.5 \\
			
			\hline
		\end{tabular}
	}
	\vskip-10pt
\end{table}

\subsection{JHU-CROWD dataset}


In this section, we discuss the benchmarking of recent algorithms including the proposed method on the new dataset. \\

\begin{table*}[htp!]
	\centering
	\caption{Results on JHU-CROWD dataset.}
	\label{tab:uccd}
	\resizebox{1\textwidth}{!}{%
		\begin{tabular}{|l|c|c|c|c|c|c|c|c|c|c|c|c|}
			\hline
			Category                              & \multicolumn{2}{c|}{Distractors} & \multicolumn{2}{c|}{Low} & \multicolumn{2}{c|}{Medium} & \multicolumn{2}{c|}{High} & \multicolumn{2}{c|}{Weather} & \multicolumn{2}{c|}{Overall} \\ \hline
			Method                                				& MAE             & MSE            & MAE         & MSE        & MAE          & MSE          & MAE         & MSE         & MAE           & MSE          & MAE           & MSE          \\ \hline
			MCNN \cite{zhang2016single}           &     103.8  &    238.5   &  37.7            &  92.5          &     84.1       &  185.2          & 499.6           &   795.5     		&     128.2           &         288.3        & 109.3     & 291.0       \\ 
			CMTL \cite{sindagi2017cnnbased}       &   135.8  &    263.8   &  47.0            &  106.0         &     82.4      &  198.3         & 407.8           &   660.2    		&     117.8           &         260.1         & 102.5      & 262.6        \\ 
			Switching CNN \cite{sam2017switching} &       100.5          &  235.5              &   32.1          & 80.5          &     76.1        &     173.1         &       395.1      &  640.1           &      105.1         &    245.2          & 99.1          & 255.1        \\						CP-CNN \cite{sindagi2017generating} &       90.5          &  210.3              &   30.2          & 71.3          &     71.2        &     155.3         &       390.2      &  620.9           &      99.9         &    243.3          & 93.5          & 238.5        \\
			SA-Net(image-based) \cite{cao2018scale} &       71.9          &  167.7              &   30.0          & 76.6         &         65.4        &       121.5      &  516.3  &   762.7      &      99.4      &    234.9          & 98.0         & 260.3       \\  
			CSR-Net \cite{li2018csrnet} &       44.3        &  102.4            &   15.8          & 39.9         &     48.4        &     77.7        &       463.5      &  746.1          &      96.5         &    284.6       & 78.4          & 242.7        \\
			CG-DRCN (proposed)                     &         \bf{43.4}        &   \bf{97.8}               &     \bf{15.7}        &   \bf{38.9}         &   \bf{44.0}           &       \bf{73.2}       &      \bf{346.2}       &       \bf{569.5}      &      \bf{80.9}        &     \bf{227.31}         & \bf{66.1}          & \bf{195.5}        \\ \hline
		\end{tabular}%
	}
	\vskip-10pt
\end{table*}

\begin{table}[htp!]
	\centering
	\caption{Results on ShanghaiTech dataset \cite{zhang2016single}.}
	\vskip-5pt
	\label{tab:shtech}
	\resizebox{\linewidth}{!}{
		\begin{tabular}{|l|c|c|c|c|}
			\hline
			& \multicolumn{2}{c|}{Part-A} & \multicolumn{2}{c|}{Part-B} \\ \hline
			Method          & MAE          & MSE          & MAE          & MSE          \\ \hline
			Cascaded-MTL \cite{sindagi2017cnnbased}           & 101.3        & 152.4        & 20.0         & 31.1         \\ 
			Switching-CNN \cite{sam2017switching}           & 90.4        & 135.0        & 21.6         & 33.4         \\ 
			CP-CNN \cite{sindagi2017generating}           & 73.6        & 106.4        & 20.1         & 30.1         \\ 
			IG-CNN \cite{babu2018divide}           & 72.5        & 118.2        & 13.6         & 21.1          \\ 
			Liu \etal \cite{liu2018leveraging}           & 73.6        & 112.0        & 13.7         & 21.4        \\
			D-ConvNet \cite{shi2018crowd_negative}           & 73.5        & 112.3        & 18.7         & 26.0           \\ 
			CSRNet \cite{li2018csrnet}           & 68.2        & 115.0        & 10.6         & 16.0    \\
			ic-CNN \cite{ranjan2018iterative}           & 69.8        & 117.3        & 10.7         & 16.0       \\
			SA-Net (image-based) \cite{cao2018scale}           & 88.1        & 134.3        & -         & -       \\
			SA-Net (patch-based) \cite{cao2018scale}           & {67.0}        & {104.5}        &  {{8.4}}         &  {{13.6}}       \\
			ACSCP  \cite{shen2018adversarial}           & 75.7        & 102.7        & 17.2         & 27.4       \\
			Jian \etal  \cite{jiang2019crowd} & {{64.2}}        & {{109.1}}        & {\bf{8.2}}         & {\bf{12.8}}\\
			CG-DRCN (proposed) & \textbf{64.0}        & \textbf{98.4}        & {8.5}         & {14.4}         \\ \hline
		\end{tabular}
	}
	
\end{table}

\begin{table}[htp!]
	\centering
	\caption{Results on UCF-QNRF datastet \cite{idrees2018composition}.}
	\label{tab:resultsucf}
	\vskip-5pt
	\resizebox{0.8\linewidth}{!}{
		\begin{tabular}{|l|c|c|}
			\hline
			Method & MAE & MSE \\			\hline
			Idrees \etal \cite{idrees2013multi}& 315.0 & 508.0 \\
			Zhang \etal \cite{zhang2015cross}& 277.0 & 426.0 \\
			CMTL \etal \cite{sindagi2017cnnbased}& 252.0 & 514.0 \\		
			Switching-CNN \cite{sam2017switching} & 228.0 & 445.0 \\
			Idrees \etal \cite{idrees2018composition}& {132.0} & {{191.0}} \\
			Jian \etal \cite{jiang2019crowd}  & {{113.0}} & {{188.0 }}\\
			CG-DRCN (proposed) & {\textbf{112.2}} & {\bf{176.3 }}\\
			\hline
		\end{tabular}
	}
\end{table}

\noindent\textbf{Benchmarking and comparison}.  We  benchmark recent algorithms on the newly proposed JHU-CROWD dataset. Specifically, we evaluate the following recent works: mulit-column network (MCNN) \cite{zhang2016single}, cascaded multi-task learning for crowd counting (CMTL) \cite{sindagi2017cnnbased}, Switching-CNN \cite{sam2017switching}, CSR-Net \cite{li2018csrnet} and SANet \cite{cao2018scale} \footnote{We used the implementation provided by \cite{gao2019c}}. Furthermore, we also evaluate the proposed method (CG-DRCN) and demonstrate  its effectiveness over the other methods. 

All the networks are trained using the entire training set and evaluated under six different categories. For a fair comparison, the same training strategy (in terms of cropping patches), as described in Section   \ref{ssec:training_details}, is used. Table \ref{tab:uccd} shows the results of the above experiments for various sub-categories of images in the test set. It can be observed that the proposed method outperforms the other methods in general. Furthermore, it may also be noted that the overall performance does not necessarily indicate the proposed method performs well in all the sub-categories. Hence, it is essential to compare the methods for each of the sub-category.

\subsection{Comparison on other datasets}

\noindent\textbf{ShanghaiTech}: The proposed network is trained on the train splits using the same strategy as discussed in Section \ref{ssec:training_details}.  Table \ref{tab:shtech} shows the results of the proposed method on ShanghaiTech  as compared with several recent approaches (\cite{sam2017switching}, \cite{sindagi2017generating}, \cite{babu2018divide}, \cite{shi2018crowd_negative},  \cite{liu2018leveraging},  \cite{li2018csrnet},  \cite{ranjan2018iterative} , \cite{cao2018scale}, \cite{shen2018adversarial} and \cite{jiang2019crowd}).  It can be observed that the proposed method outperforms all existing methods on Part A of the dataset, while achieving comparable performance on Part B. \\

\noindent\textbf{UCF-QNRF}: Results on the  UCF-QNRF \cite{idrees2018composition} dataset as compared with recent methods ( \cite{idrees2013multi},\cite{zhang2016single},\cite{sindagi2017cnnbased}) are shown in Table \ref{tab:resultsucf}. The proposed method is compared against  different approaches: \cite{idrees2013multi},   \cite{zhang2016single},   \cite{sindagi2017cnnbased},\cite{sam2017switching},  \cite{idrees2018composition} and \cite{jiang2019crowd}. It can be observed that the proposed method outperforms other methods by a considerable margin.

\section{Conclusions}
In this paper, we presented a novel crowd counting network that employs residual learning mechanism in a progressive fashion to estimate coarse to fine density maps. The efficacy of residual learning is further improved by introducing an uncertainty-based confidence weighting     mechanism that is designed to enable the network to propagate only high-confident residuals to the output. Experiments on recent datasets demonstrate the effectiveness of the proposed approach.  Furthermore, we also introduced a new large scale unconstrained crowd counting dataset (JHU-CROWD) consisting of 4,250 images with 1.11 million annotations. The new dataset is collected under a variety of conditions and includes images with weather-based degradations and other distractors. Additionally, the dataset provides a rich set of annotations such as head locations, blur-level, occlusion-level, size-level and other image-level labels. 

\section*{Acknowledgment}
\noindent This work was supported by the NSF grant 1910141.

{\small
	\bibliographystyle{ieee_fullname}
	\bibliography{egbib}
}

\end{document}